\documentclass{article}

\usepackage{arxiv}

\usepackage[bookmarks=false]{hyperref}
    \hypersetup{colorlinks,
      linkcolor=blue,
      citecolor=blue,
      urlcolor=blue}
\usepackage{xcolor}

\usepackage{booktabs}
\usepackage{xspace}
\usepackage{amsmath}
\usepackage{hyperref}
\usepackage{graphicx}
\usepackage{algorithm}
\usepackage{algpseudocode}
\usepackage{booktabs}
\usepackage{listings}
\usepackage{geometry}
\usepackage{siunitx}
\usepackage{bm}
\usepackage{amssymb}
\usepackage[nolist]{acronym}
\begin{acronym}[UML]
    \acro{KG}{knowledge graph}
    \acro{KGE}{Knowledge Graph Embedding}
    \acro{LLM}{Pretrained Large Language Model}
    \acro{RAG}{Retrieval-augmented Generation}
    \acro{ICR}{Interval Coverage Rate}
    \acro{IW}{Interval Width}
\end{acronym}

\newcommand{\approach}{\textsc{RALP}\xspace}
\newcommand{\learnapproach}{\ensuremath{\textsc{RALP}^\dagger}\xspace}

\newcommand{\entities}{\ensuremath{\mathcal{E}}\xspace}
\newcommand{\relations}{\ensuremath{\mathcal{R}}\xspace}
\newcommand{\kg}{\ensuremath{\mathcal{G}}\xspace}
\newcommand{\scoreFunc}{\phi}
\newcommand{\triple}[3]{\ensuremath{(#1, #2, #3)}}

\newcommand{\pair}[2]{\ensuremath{(#1, #2)}}

\newcommand{\stringparameters}{\ensuremath{\beta^{\leq c}}}

\graphicspath{ {./images/} }

\title{Learning Chain Of Thoughts Prompts for Predicting Entities,
Relations, and even Literals on Knowledge Graphs}

\author{
 Alkid Baci, \ Luke Friedrichs, \ Caglar Demir, \ N'Dah Jean Kouagou, \ Axel-Cyrille Ngonga Ngomo \\
  Department of Computer Science \\
  Paderborn University \\
  Warburger Str. 100, 33098 Paderborn, Germany \\
  \texttt{\{alkid.baci, caglar.demir, ndah.jean.kouagou, axel.ngonga\}@upb.de} \\
  \texttt{lukef@mail.uni-paderborn.de} \\
}

\begin{document}
\maketitle
\begin{abstract}
Knowledge graph embedding (KGE) models perform well on link prediction but struggle with unseen entities, relations, and especially literals, limiting their use in dynamic, heterogeneous graphs. In contrast, pretrained large language models (LLMs) generalize effectively through prompting. We reformulate link prediction as a prompt learning problem and introduce \approach, which learns string-based chain-of-thought (CoT) prompts as scoring functions for triples. Using Bayesian Optimization through MIPRO algorithm, \approach identifies effective prompts from fewer than 30 training examples without gradient access.
At inference, \approach predicts missing entities, relations or whole triples and assigns confidence scores based on the learned prompt. We evaluate on transductive, numerical, and OWL instance retrieval benchmarks. \approach improves state-of-the-art KGE models by over 5\% MRR across datasets and enhances generalization via high-quality inferred triples. On OWL reasoning tasks with complex class expressions (e.g., $\exists hasChild.Female$, $\geq 5 \; hasChild.Female$), it achieves over 88\% Jaccard similarity. These results highlight prompt-based LLM reasoning as a flexible alternative to embedding-based methods. We release our implementation, training, and evaluation pipeline as open source: https://github.com/dice-group/RALP.
\end{abstract}

\section{Introduction}

Knowledge graphs (KGs) are a foundational component of modern AI systems, supporting applications such as search~\cite{wang2019knowledge, xiong2020approximate}, question answering~\cite{edge2024local, baek-etal-2023-knowledge, he2024g}, and recommender~\cite{guo2020survey, DBLP:journals/corr/abs-1904-12575} systems. Central to maintaining their utility is the task of link prediction—inferring missing facts by modeling the underlying structure of the graph. Traditional knowledge graph embedding (KGE) models~\cite{dettmers2018convolutional, ruffinelli2020you, dai2020survey} have made progress toward link prediction problem by representing entities and relations as continuous vectors and applying learned scoring functions. However, these methods are inherently limited when faced with unseen entities or relations, and they struggle to reason over literals, such as numerical values or descriptive text, which are pervasive in real-world KGs.

Meanwhile, LLMs have demonstrated remarkable capabilities in few-shot and zero-shot learning through prompting—conditioning the model on examples or task instructions \cite{kojima2022large}. Prompting plays a critical role in retrieval-augmented generation (RAG) systems \cite{liu2023pre}, where relevant context from external sources is injected into prompts to improve task performance~\cite{lewis2020retrieval, ram2023context, petroni-etal-2021-kilt}.

While many RAG pipelines rely on handcrafted prompt templates, recent work explores automated prompt optimization using techniques such as APE~\cite{zhou2022large}, OPRO~\cite{yang2024large}, EvoPrompt~\cite{guo2024connecting}, and MIPRO~\cite{opsahl-ong-etal-2024-optimizing}. DSPy~\cite{khattab2024dspy} further formalizes LLM pipelines as transformation graphs with learnable prompts via declarative modules. These advances suggest that high-quality prompts can be learned to maximize downstream performance—even without access to gradients.

In this work, we revisit the classic link prediction problem in knowledge graphs and propose a new formulation: learning a string-parametrized scoring function $\scoreFunc_{\beta^{\leq c}}$ instead of the conventional continuous vector-parametrized one $\scoreFunc_{\Theta}$. Traditional knowledge graph embedding models~\cite{dettmers2018convolutional, ruffinelli2020you, dai2020survey} represent entities and relations as dense vectors and compute a real-valued score $\hat{y} = \scoreFunc(h, r, t)$ to estimate triple plausibility. In contrast, our method learns a prompt $\beta^{\leq c}$—a CoT string of up to $c$ tokens—that conditions an LLM to score triples, including those with unseen entities, relations, or literals. To our knowledge we are the first approach to treat numerical link prediction via prompt-based reasoning, using LLMs directly without regression layers or binning and achieving promising results.

\approach key advantages:
\begin{itemize}
\item[$\checkmark$] Requires only a few examples ($\leq$ 30) to optimize $\beta^{\leq c}$ via gradient-free MIPRO;
\item[$\checkmark$] Supports reasoning over unseen entities, relations, and literals;
\item[$\checkmark$] Augments training data to improve downstream KGE models;
\item[$\times$] Assumes that entity/relation strings are semantically meaningful.
\end{itemize}

\begin{figure*}
    \centering
    \small
    \includegraphics[width=\textwidth]{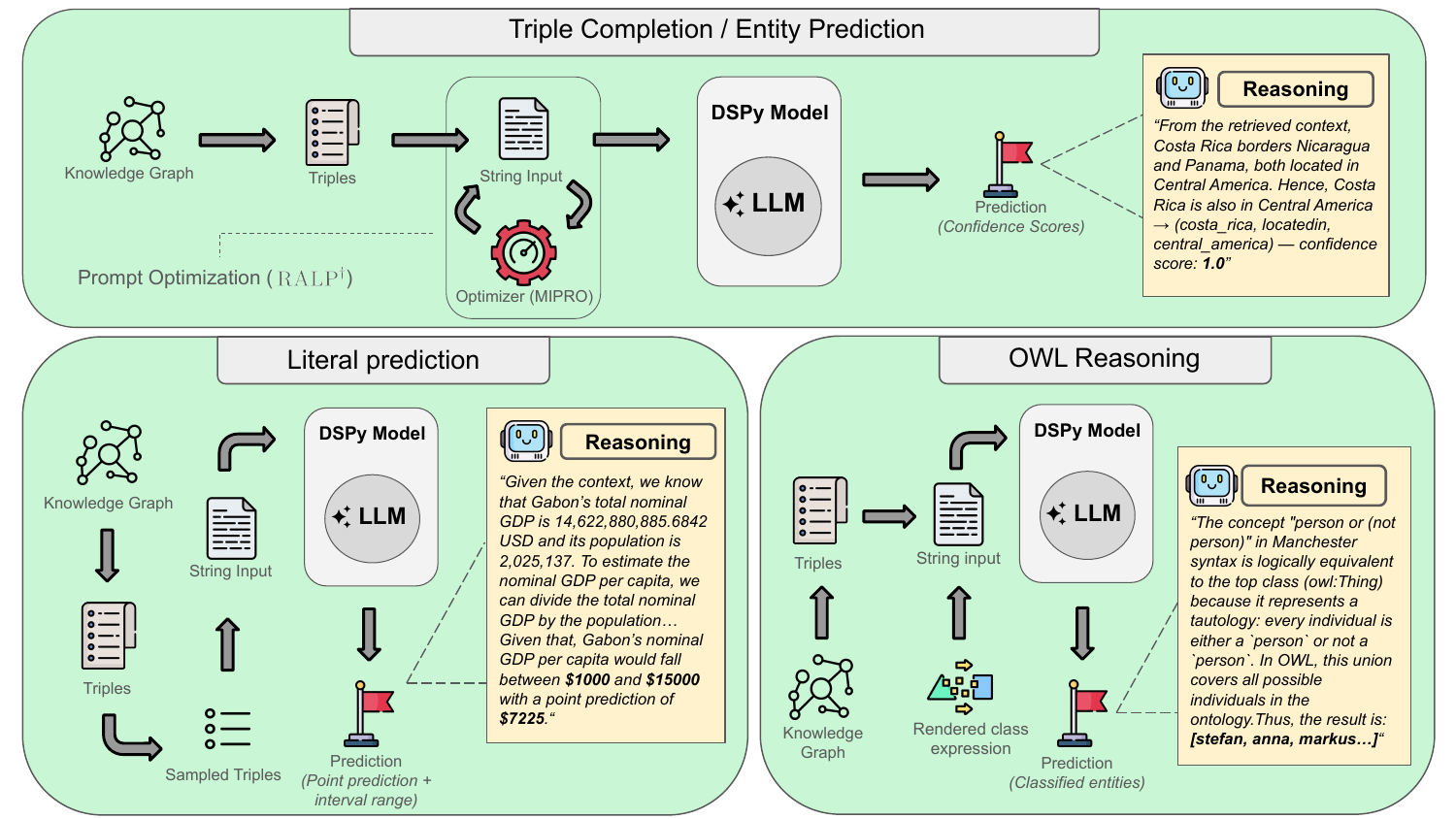}
    \caption{Overview of the RALP framework in three different tasks within the link prediction problem. Notice that sampling of triples or MIPRO optimization can be performed in all three of them but is omitted because of space limitations.} 
    \label{fig:overview}
\end{figure*}

In summary, we introduce \approach, a method that shifts the perspective of solving the link prediction problem by learning symbolic, CoT-based scoring functions over KGs using LLMs and automated prompt tuning. Our method generalizes beyond traditional KGE models, achieving strong performance across transductive, numerical, and OWL instance retrieval benchmarks—paving the way for symbolic reasoning with LLMs over rich and dynamic knowledge graphs. An overview of the \approach framework is given in Figure \ref{fig:overview}.

\section{Background and Related Work}
\label{sec:background}

\subsection{Knowledge Graphs}
A knowledge graph represents structured collections of assertions describing the world~\cite{hogan2021knowledge}.
These collections of assertions have been used in a wide range of applications, including drug discovery, web search, recommendation and question answering~\cite{hogan2021knowledge}
Formally, a \ac{KG} is often defined as a set of triples $\kg:= \{ \triple{h}{r}{t}\in \entities \times \relations \times \entities\}$, where \entities\ and \relations\ stand for a set of entities and a set of relations, respectively~\cite{dettmers2018convolutional,demir2023clifford}. 
Each triple $\triple{h}{r}{t} \in \kg$  represents an  assertion based on two entities $h, t \in \entities$ and a relation $r \in \relations$.
This formulation can be further extended to generalize over triples involving literals, e.g.,
$\kg:= \{ \triple{h}{r}{t}\in \entities \times \relations \times \entities \cup \mathcal{I} \}$, where $\mathcal{I}$ denotes the set of unique numerical literals.
These triples can be inferred from an existing set of triples by means of designing logical rules or learning continuous vector representations via knowledge graph embedding models~\cite{hogan2021knowledge}.

\subsection{Knowledge Graph Embeddings and Link Prediction}
Most KGE models map entities $e \in \mathcal{E}$ 
and relations $r \in \mathcal{R}$ found in a Knowledge Graph (KG) $\kg \subseteq \mathcal{E} \times \mathcal{R} \times \mathcal{E}$ to $\mathbb{V}$, where $\mathbb{V}$ is a $d$-dimensional vector space and $d \in \mathbb{N} \backslash \{0\}$~\cite{demir2023clifford}. 
Most KGE models are often designed as a scoring function to 
learn continuous vector representations tailored towards link prediction~\cite{hogan2021knowledge}.
They are often defined as parameterized scoring functions $\scoreFunc_\Theta: \entities \times \relations \times \entities \mapsto \mathbb{R}$,
where $\Theta$ denotes parameters and often comprise 
entity embeddings $\mathbf{E} \in \mathbb{V}^{|\mathcal{E}| \times d_e}$, relation embeddings $\mathbf{R} \in \mathbb{V}^{|\mathcal{R}| \times d_r}$, and additional parameters.
Given $\triple{h}{r}{t} \in \entities \times \relations \times \entities$, the prediction $\hat{y}:=\scoreFunc_\Theta\triple{h}{r}{t}$ signals the likelihood of $\triple{h}{r}{t}$ being true~\cite{dettmers2018convolutional,demir2023clifford}.
Since $\kg_{\text{train}}$ contains only assertions that are assumed to be true, assertions assumed to be false are often generated by applying the negative sampling, 1vsAll or Kvsall training strategies~\cite{ruffinelli2020you}.
Regardless of the selected training strategies, the training is often realized in the mini-batch stochastic gradient descent or its sophisticated variants fashion, e.g. ADAM.
Let $\Theta$ denote $\{\mathbf{E}, \mathbf{R} \}$ randomly initialized embeddings.
An embedding update over a mini batch $\mathcal{B}$ can be defined as:
\begin{equation}
\label{eq:kge_param_update}
\Theta \leftarrow \Theta - \nabla \, \eta \, \frac{1}{|\mathcal{B}|} \sum_{x,y \in \mathcal{B}} \nabla_\Theta \mathcal{L}(x,y, \Theta)    
\end{equation}
where 
\begin{itemize}
  \item $\Theta$ are the learnable parameters of $\scoreFunc$,
  \item $\eta$ is the learning rate,
  \item $\mathcal{L}(x,y, \Theta)$ is the loss function evaluated at point $x,y$,
  \item $\nabla_\Theta \mathcal{L}(x,y, \Theta)$ is the gradient of the loss w.r.t. parameters,
  \item $\mathcal{B}$ is a mini-batch of training samples, and
  \item $\nabla$ represents the gradient operator.
\end{itemize}
The cross-entropy loss function is often used as
the $\mathcal{L}$ that is defined as

\begin{equation}
\mathcal{L}(x,y,\Theta) = - y \text{log}(\hat{y}) - (1-y) \text{log}(1-\hat{y}),
\label{eq:ce}
\end{equation}
where $\hat{y}:= \sigma(\scoreFunc_\Theta (x))$ s.t.
$x:=\triple{h}{r}{t}$ and $\sigma(\cdot)$ denotes the logistic sigmoid function $\sigma(s) = \frac{1}{1 + \text{exp}(-s)}$.
Depending on the selected training technique, the learning problem (as well as $\mathcal{B}$) is rendered as binary classification (e.g. Negative Sampling), a multi-class classification (1vsAll) or a multi-label classification (KvsAll).
The resulting models are then evaluated w.r.t. their ability of predicting missing entity rankings~\cite{ruffinelli2020you}. 
Regardless of the selected training strategies, the training is often realized in the mini-batch stochastic gradient descent or its sophisticated variants fashion.

\subsection{LLMs for Knowledge Graph Completion}

Recent work explores leveraging pretrained large language models (LLMs) for KG completion. 
Some approaches adapt model architectures or fine-tune on textualized triples. 
KGLM~\cite{youn2022kglm} incorporates entity and relation embeddings into pretrained language models via continued pretraining. 
KG-LLM~\cite{shu2024knowledge} fine-tunes LLMs using ranking-based objectives over textualized triples and employs retrieval for scalable candidate pruning. 
LINKGPT~\cite{he2024linkgpt} combines structural encodings with instruction tuning and retrieval–rerank mechanisms to improve efficiency. In contrast, prompt-based methods avoid architectural changes and gradient-based fine-tuning. 
PDKGC~\cite{geng2025prompting} formulates link prediction as multiple-choice question answering and selects prompts via ranking-guided sampling, enabling few-shot generalization.
These approaches demonstrate that LLMs can internalize graph structure when appropriately adapted. However, most rely on fine-tuning, ranking losses, or handcrafted prompt templates, and typically focus on entity prediction in transductive benchmarks.

\subsection{Automated Prompt Optimization}

Automated prompt optimization seeks to replace manual prompt engineering with principled search strategies. 
Prior work includes discrete and evolutionary search~\cite{pryzant2023automaticpromptoptimizationgradient, guo2025evopromptconnectingllmsevolutionary}, reinforcement learning and self-improvement mechanisms~\cite{huang2022largelanguagemodelsselfimprove, hu2024enablingintelligentinteractionsagent}, and large language models that optimize instructions via meta-prompting (e.g., OPRO~\cite{yang2024large}). 
DSPy~\cite{khattab2024dspy} formalizes LLM pipelines as declarative programs with learnable prompt modules. 
MIPRO~\cite{opsahl-ong-etal-2024-optimizing} introduces Bayesian optimization for jointly refining instructions and demonstrations across multi-stage pipelines, enabling gradient-free optimization under weak supervision.
These developments suggest that prompts can serve as learnable, discrete parameters optimized directly for downstream objectives.

\paragraph{Positioning of Our Work.}
Unlike embedding-based KGE models, we replace continuous vector-parameterized scoring functions $\scoreFunc_\Theta$ with a string-parameterized scoring function $\scoreFunc_{\beta^{\leq c}}$, where $\beta^{\leq c}$ is a learned chain-of-thought prompt. 
Unlike prior LLM-based KG completion methods, we require no fine-tuning, architectural modification, or ranking loss. 
Instead, we cast link prediction—including numerical and OWL-style reasoning—as a discrete prompt optimization problem solved via Bayesian optimization. 
This formulation enables reasoning over unseen entities, relations, and literals while remaining fully gradient-free and model-agnostic.

\section{Methodology}
\label{sec:methodology}

Most KGE models are defined as a scoring function $\scoreFunc_\Theta: \entities \times \relations \times \entities \mapsto \mathbb{R}$ parameterized by a vector $\Theta$.
The learning problem is often formalized as a classification problem, where the performance is measured by a cross entropy loss function.
Here, we propose a \textbf{string-parameterized scoring function} $\scoreFunc_{\beta^{\leq c}}: \entities \times \relations \times \entities \mapsto \mathbb{R}$. 
This function predicts the likelihood of a triple being true, where the parameter $\beta^{\leq c}$ is a sequence of at most $c$ vocabulary terms (tokens) $\beta_i \in \Sigma$. Specifically, $\beta^{\leq c}$ represents the input prompt provided to a selected pretrained large language model \texttt{LM}, and $c$ denotes the context length of \texttt{LM}.
The prompt $\beta^{\leq c}$ is constructed for a given query input (e.g. pair $\pair{s}{p} \in \entities \times \relations$ or a triple $\triple{s}{p}{o} \in \entities \times \relations \times \entities$) by a function $\Pi(\cdot)$. 
$\Pi(\cdot)$ can be interpreted as an initialization scheme for $\Theta$, similar to how neural networks use random uniform or Xavier initialization \cite{glorot2010understanding}.
\textbf{Our goal is to learn such $\beta^{\leq c}$ initialized by $\Pi(\cdot)$ to accurately predict the likelihoods of triples being elements of $\mathcal{G}$ into $\beta^{\leq c}$, while maximizing the link prediction performance}.
Therefore, we aim to learn $\beta^{\leq c}$ tailored towards the link prediction task.
\subsection{Learning Chain of Thoughts Prompts for Link Prediction}
To find suitable $\scoreFunc_{\beta^{\leq c}}$ for a given $\mathcal{G}$ and a selected $\texttt{LLM}$, we propose to use 
a gradient-free search method (MIPRO).
As the guiding signal, we use the cross entropy as performance metric (e.g. MRR scores) averaged over an evaluation set, 
rather than a loss gradient averaged over a training mini-batch.
While MIPRO utilizes Bayesian Optimization over a discrete search space of possible strings rather than gradient descent on continuous parameters $\Theta$, its process is still iterative and data-driven, aiming to find an optimal string configuration. The ``update'' in MIPRO refers to the process by which the optimizer refines its search strategy and identifies the best-performing string parameter found so far based on evaluation results.

Let $\beta^{\leq c}$ represent a specific candidate string parameter, a sequence of at most $c$ vocabulary terms $\beta_i \in \Sigma$. 
$\Sigma$ corresponds to the vocabulary size of the byte pair encoded tokens. \footnote{$|\Sigma|=151936$ 
when using the Qwen2.5 tokenizer~\cite{yang2024qwen2}.}
The set of all possible candidate string parameters forms a discrete search space $\mathcal{S}_\Sigma$. Our objective is to find $\beta^{*\leq c} \in \mathcal{S}_\Sigma$ that maximizes a performance metric on a validation set, such as the cross-entropy loss $\mathcal{L}$ averaged over evaluation triples.
MIPRO performs a sequence of $T$ trials. At each trial $t$, the Bayesian Optimization algorithm does the following:

\begin{enumerate}
    \item \textbf{Select Candidate $\beta^{\leq c}_t$:} Based on the observed performance of string parameters evaluated in previous trials $$\{ (\beta^{\leq c}_i, Q(\beta^{\leq c}_i)) \}_{i=1}^{t-1},$$
    a probabilistic model of the $Q$ function over $\mathcal{S}_\Sigma$ is updated. An   function 
    $$a(\beta^{\leq c} | \{ (\beta^{\leq c}_i, Q(\beta^{\leq c}_i)) \}_{i=1}^{t-1})$$
    is then used to select the next candidate string parameter $\beta^{\leq c}_t$ that is most promising for evaluation (e.g., maximizing expected improvement)
    $$ \beta^{\leq c}_t = \arg\max_{\beta^{\leq c} \in \mathcal{S}_\Sigma} a(\beta^{\leq c} | \{ (\beta^{\leq c}_i, Q(\beta^{\leq c}_i)) \}_{i=1}^{t-1}).$$
    This step is the analogue to the ``update'' step in gradient descent elucidated in Equation \ref{eq:kge_param_update}, as it determines the next point in the search space ($\mathcal{S}_\Sigma$) to explore, but it is based on probabilistic modeling and acquisition rather than a gradient calculation on continuous parameters.

    \item \textbf{Evaluate $Q(\beta^{\leq c}_t)$:} The selected candidate string parameter $\beta^{\leq c}_t$ is used with the scoring function $\scoreFunc_{\beta^{\leq c}_t}$ and evaluated on a (mini)batch of the validation set to obtain its score.
    \begin{equation}
        s_t = Q(\beta^{\leq c}_t)
    \end{equation}

    \item \textbf{Update Best Found String Parameter:} The best string parameter found so far, $\beta^{*\leq c}_{t}$, is updated if the current candidate $\beta^{\leq c}_t$ yields a higher score than the best score recorded in previous trials.
    \begin{equation}
    \beta^{*\leq c}_{t} = \begin{cases} \beta^{\leq c}_t & \text{if } s_t > Q(\beta^{*\leq c}_{t-1}) \\ \beta^{*\leq c}_{t-1} & \text{otherwise} \end{cases}        
    \end{equation}
    where $\beta^{*\leq c}_0$ is the initial string parameter generated by $\Pi(\cdot)$.
\end{enumerate}

After $T$ trials, the optimization process terminates, and the final best-found string parameter $\beta^{*\leq c}_T$ is returned. Thus, the string parameter update involves the iterative selection of candidate string parameters guided by the Bayesian Optimization model and the tracking of the single best-performing string parameter found over the search process, rather than incrementally modifying continuous parameters like $\Theta$. 
The learned $\beta^{*\leq c}_T$ then serves as the optimized parameter for the scoring function $\scoreFunc_{\beta^{*\leq c}_T}$.
\subsubsection{Initialization:}
$\Pi(\cdot)$ combines relevant information about the query, a definition of the link prediction task, and few-shot examples $\mathcal{D}_{\text{few}}$. $\mathcal{D}_{\text{few}}$ is a set of query-answer pairs illustrating the task
$\mathcal{D}_{\text{few}}$ can be defined through negative sampling technique as well as KvsAll training
$ \mathcal{D}_{\text{few}} = \{ [ \pair{s'}{r'}, \{ o_{1}, \ldots, o_{n} \} ] \}$
,e.g.
the following holds in the KvsAll setup: $\forall o_i \in \{ o_{1}, \ldots, o_{n} \}$  $\triple{s'}{r'}{o_i} \in \mathcal{G}$.
Thus, the prompt is formulated as:
\begin{equation}
\beta^{\leq c} = \Pi(\mathcal{D}_{\text{few}}, s, r).
\end{equation}
$\beta^{\leq c}$ and $\Pi$ can be predefined through expert prompt engineering or learned to maximize a downstream metric without access to gradients.
In this work, we use the cross entropy loss function as $Q$ defined as
\begin{equation}
    Q(\beta^{\leq c}) = \frac{1}{|\mathcal{B}|} \sum_{x,y \in \mathcal{B}}  - y \text{log}(\hat{y}) - (1-y) \text{log}(1-\hat{y}),
\end{equation}
where $y \in \{0,1 \}$ and $\hat{y}:= \sigma(\scoreFunc_{\beta^{\leq c}} (x))$ s.t. $x:=\triple{h}{r}{t}$ and $\sigma(\cdot)$ denotes the logistic sigmoid function $\sigma(s) = \frac{1}{1 + \text{exp}(-\mathbf{s})}$.

\subsubsection{Approaches:}
Here, we propose two main approaches:
\approach  and \learnapproach.
While \approach directly utilizes the in-context learning ability of the selected \texttt{LM} with a default prompt,
\learnapproach uses the MIPRO algorithm to learn 
$\beta^{\leq c}$ tailored towards the link prediction task on a given Knowledge Graph (KG) $\mathcal{G}$.
Given a subject entity and a relation $(s, r)$, \approach predicts missing information (e.g., candidate objects for a $(s, r, ?)$ query) and potentially their likelihoods by processing the prompt:
\begin{equation}
\hat{\mathbf{y}}=:\approach(s, r) = \texttt{LM}(\Pi(\mathcal{D}_{\text{few}}, s, r)),
\end{equation}
where $\hat{\mathbf{y}} \in [0,1]^{|\entities|}$ predicted probabilites for entities, \pair{s}{r} denoting input pair, and \texttt{LM} denoting a selected pretrained LLM.
The output of \texttt{LM} is then parsed to obtain predictions. \learnapproach uses MIPRO to learn a prompt specifically for the link prediction task as described above.
\approach can be readily used to predict missing triples as well as its predictive capability can be leveraged to enrich the training data. We propose Algorithm \ref{alg:enrichment} to identify potential missing triples within the training graph by querying \approach for existing head-relation pairs and adding highly-scored, non-existent triples to a set of missing triples.

\begin{algorithm}[H]
\caption{Knowledge Graph Enrichment via In-context Learning}
\label{alg:enrichment}
\begin{algorithmic}[1]
\Require \approach: Link prediction function using LLM inference; $\mathcal{G}_{\text{train}}$: Training triples; $\theta$: Confidence threshold ($> 0.5$)
\State Initialize $\mathcal{G}_{\text{missing}} \gets \emptyset$
\ForAll{$(s, p, o) \in \mathcal{G}_{\text{train}}$}
\State // Query \approach for potential objects given $(s, p)$
\State Let $\mathcal{S}$ be the set of predicted objects and their scores obtained from $\approach(s, p)$
\ForAll{$o'$ predicted by $\approach(s, p)$ with score $\mathcal{S}[o'] > \theta$}
\If{$(s, p, o') \notin \mathcal{G}_{\text{train}}$}
\State $\mathcal{G}_{\text{missing}} \gets \mathcal{G}_{\text{missing}} \cup \{(s, p, o')\}$
\EndIf
\EndFor
\EndFor
\State \Return $\mathcal{G}_{\text{missing}}$
\end{algorithmic}
\end{algorithm}
Algorithm \ref{alg:enrichment} iterates through every subject-relation pair present in the training graph (derived from triples $(s, p, o) \in \mathcal{G}_{\text{train}}$). For each pair $(s, p)$, it queries \approach, which predicts a set of potential object entities $\mathcal{S}$ along with confidence scores. If a predicted object $o'$ for $(s, p)$ has a score exceeding the threshold $\theta$ and the triple $(s, p, o')$ does not already exist in the training graph, it is considered a potentially missing true triple and added to $\mathcal{G}_{\text{missing}}$.
In one of our experiments, we augment the training splits of benchmark datasets with
respective $\mathcal{G}_{\text{missing}}$ to measure the benefits of using \approach or \learnapproach as an augmentation technique for conventional \ac{KGE} models.

\subsection{Numerical Literal Prediction}
\label{label:numerical_prediction}

For numerical prediction, $Q(\cdot)$ can be defined as a regression loss (e.g., MSE). 
Given $(s,r)$ where $r$ is a data property, prediction proceeds via contextual retrieval. We construct contextual subsets:
$\mathcal{G}_{(s)} = \left\{ (s', r', t) \in \mathcal{G} \,\middle|\, s' = s \ \wedge \ r' \neq r \ \right\}$ and 
$
\mathcal{G}_{(r)} = \left\{ (s', r', t) \in \mathcal{G} \,\middle|\, r' = r \wedge \ s' \neq s \right\}$ where
$\mathcal{G}_{(s)}$ provides subject-specific context, while $\mathcal{G}_{(r)}$ supplies relation-level value patterns. 
Due to token constraints, subsets are sampled with priority given to $\mathcal{G}_{(s)}$ to facilitate chain-of-thought reasoning. 
The resulting few-shot context is passed to $\approach$ to predict the numerical literal.

\section{Experimental Setup}
\subsection{Datasets}
We utilized the benchmark datasets Countries-S1, Countries-S2, Countries-S3, LitWD1K, and Father \cite{demir2022hardware,gesese_2021_4701190,demir2025ontolearn}.
These datasets are consider relatively small but nevertheless they effectively show the capabilities of our approach. While applying our approach to larger datasets is limited by the token size that the LLM can process, we put focus in this paper on highlighting the novelty of our approach while leaving scalability to future work. The LitWD1K dataset, derived from Wikidata, includes numerical literals. For our experiments, we focused on two subsets: numeric\_literals (10988 triples of the form \triple{s}{r}{l} where l is a numerical literal) and the 'train' set ($26115$ triples of the form (s,r,o)). To improve the understanding of the LM, we preprocessed LitWD1K by replacing Wikidata IDs with their corresponding labels, creating two additional batches. 
We specifically considered only float/integer literals within the numeric literals set.
\subsection{Transductive Link Prediction Setup}
We used the best hyperparameter reported in the dice-framework \cite{demir2022hardware}.
Throughout our experiments, each entity and relation is represented with 32-dimensional real valued vector across datasets and models, e.g., TransE, DistMult, MuRE in $\mathbb{R}^{32}$ 
and ComplEx, Keci, DeCaL in $\mathbb{C}^{16}$,
and QMult, QuatE $\mathbb{H}^8$.
We use the Adam optimizer with 0.1 learning rate and train each model for 256 epochs with the batch size of 1024.
The dropout rate on the embeddings is set to 0.3.

\subsection{Numerical Literal Link Prediction Setup}

Given a \pair{s}{r} pair from the \textit{numeric\_literals} set, we retrieve the relevant context for this pair and generate 3 values,
$\hat{y}, \ \hat{y}_{\text{min}}$ and $\hat{y}_{\text{max}}$. Each of these values, as well as $y$, the true value of $l$ for this pair, is stored in 2-dimensional arrays. Due to large number of properties, we select a subset $\mathcal{R}' \subseteq \mathcal{R}_{\text{nc}}$  uniformly at random, where $\mathcal{R}_{\text{nc}}$ denotes the set of all unique properties in \textit{numeric\_literals} batch and $|\mathcal{R}'| = 10$. We proceed to create predictions for each \pair{s}{r} where $r \in \mathcal{R}'$ and store the values grouped by each unique $r$.
For our evaluation metric, we compute the mean value for the array that stores the $y$ values and the one that stores the $\hat{y}$ values. We also show the standard deviation for the $y$ values. Whereas for checking the calibration of the interval $[\hat{y}_{\text{min}}, \hat{y}_{\text{max}}]$ we compute interval coverage rate (IRC):
\begin{equation}
    \text{ICR} = \frac{1}{N} \sum_{i=1}^{N} \mathbf{1} \left( \hat{y}_{\text{min}}^{(i)} \leq y^{(i)} \leq \hat{y}_{\text{max}}^{(i)} \right),
\end{equation}

where $N$ is the total number of triples where the given property appears. To evaluat the sharpness of the interval we also compute the interval width (IW):

\begin{equation}
    \text{IW} = \frac{1}{N} \sum_{i=1}^{N} \left( \hat{y}_{\text{max}}^{(i)} - \hat{y}_{\text{min}}^{(i)} \right).
\end{equation}

\subsection{Experimental Setup OWL Reasoning}
To evaluate the ability of large language models (LLMs) to predict instances of a given concept, we employed a two-step pipeline on the \textit{Father} dataset, which contains 130 concept expressions formulated in the $\mathcal{ALCQHI}$ description logic.

The signature of the few-shot generation is defined as follows:
\textbf{Input fields:} $\mathcal{G}$, $C$ (concept expression in $\mathcal{L}_{\mathcal{ALCQHI}}$), $S$ (LLM-generated explanation of $C$'s syntax), $T \subseteq \mathcal{E}$ (ground truth set of true entities satisfying $C$).
\textbf{Output field:} $E$ (CoT example demonstrating inference of $T$ from $C$ using $\mathcal{G}$).

We decided to add the syntax explanation $S$ because it proved beneficial in reducing hallucinations and guiding accurate logical reasoning during empirical testing.

The second step involves a \texttt{ChainOfThought} DSPy module, which predicts the set of instances corresponding to the input concept expression. The module is defined with the following signature:
\textbf{Input fields:}
$G$: The knowledge graph as above,
$C$: The target concept expression,
$\mathcal{E}_{\text{examples}}$: Few-shot CoT examples generated from Step 1.

\textbf{Output field:}
$\hat{T}$: The predicted set of entities that are classified as instances of concept $C$.

To assess the performance of the model, we compute the \textit{Jaccard similarity} between the predicted and true instance sets.
Concepts are grouped according to their logical or structural characteristics as detailed in Table~\ref{tab:js_per_concept_type}, and we report the mean Jaccard similarity score for each group.
In total, two few-shot examples are generated using two distinct concept expressions along with their corresponding ground truth sets.
\subsection{Hardware and LLM Setup}
Throughout our experiments, we use an Ubuntu server with Intel(R) Xeon(R) Platinum with 4 NVIDIA H100 80GBs. Thereon, we locally hosted a vLLM instance of Qwen/Qwen2.5-32B-Instruct with 4 tensor core parallelization~\cite{kwon2023efficient,qwen2.5,qwen2}.

\section{Results}
\subsection{Transductive Link Prediction}
\label{sec:transduictive_lp}
Table \ref{table:countries_with_llarp} report the impact of MIPRO optimization in the link prediction performances. 
Results suggest that \learnapproach even MIPRO light optimization reaches a new state-of-the-art link prediction performances on the Countries dataset.
To best of our knowledge, there is no state-of-the-art knowledge graph embedding model reaching 1.000 MRR or Hits@1 on the S1 dataset.
During optimization, \learnapproach with MIPRO medium learned the following two prompts:
\begin{quote}
    \textbf{Learned Composer Prompt}: \textit{"You are a geography expert. Given a subject and a predicate, your task is to find multiple tail entities that could be related to the subject through the predicate. For example, if the subject is a country and the predicate is 'locatedin', you should provide the continents or regions where the country is located. If the predicate is 'neighbor', you should provide the names of the countries that border the subject country. Your response should include reasoning that explains the thought process leading to the prediction and a list of candidate entities found among the provided entities."}
\end{quote}
and
\begin{quote}
    \textbf{Learned Scorer Prompt}:
    \textit{"Given a list of predicted entities, use your understanding of geographical relationships and historical contexts to provide reasoning and likelihood scores for each entity. Consider the broadness or specificity of the entity, as well as any known connections or conflicts within the region. Your reasoning should explain why each entity has been assigned its respective score."}
\end{quote}
\begin{table*}
    \caption{
    Link prediction results of \learnapproach on Countries-S1, Countries-S2, and Countries-S3. \approach single
    Bold and underlined results indicate the best results and second best results. } 
    \centering
    \small
    \setlength{\tabcolsep}{5pt}
    \begin{tabular}{l cccc cc cccc cc cccc}
    \toprule
     \textbf{Models} &\multicolumn{4}{c}{\textbf{S1}} && \multicolumn{4}{c}{\textbf{S2}}&& \multicolumn{4}{c}{\textbf{S3}}\\
     \cmidrule(l){2-5} \cmidrule(l){7-10} \cmidrule(l){12-15}
        &MRR  &@1   &@3   &@10                                       && MRR & @1  & @3  & @10  && MRR & @1  & @3  & @10\\
    \midrule
    \learnapproach light& 0.979 & 0.958 & \textbf{1.000} & \textbf{1.000}  && 0.750 & 0.500 & \textbf{1.000} & \textbf{1.000} && 0.595 & 0.375 & 0.791 & 0.833\\
    \learnapproach medium & \textbf{1.000} & \textbf{1.000} & \textbf{1.000} & \textbf{1.000} && 0.750 & 0.500 & \textbf{1.000} & \textbf{1.000}
    && \textbf{0.764} & \textbf{0.541} & \textbf{1.000} & \textbf{1.000}\\
     \bottomrule
    \end{tabular}
    \label{table:countries_with_llarp}
\end{table*}
Table~\ref{table:countries_enriched} report the link prediction performance of state-of-the-art embeddings models w/o \learnapproach. 
Results suggest that \learnapproach can be used as an effective data augmenter to learn better representations.
\begin{table*}
    \caption{
    Link prediction results on Countries-S1, Countries-S2, and Countries-S3.
    Link prediction results on the training and testing datasets are denoted with train and test.
    Bold results indicate the best generalization performance of a respective model.}
    \centering
    \small
    \setlength{\tabcolsep}{5pt}
    \begin{tabular}{l cccc cc cccc cc cccc}
    \toprule
     \textbf{Models} &\multicolumn{4}{c}{\textbf{S1}} && \multicolumn{4}{c}{\textbf{S2}}&& \multicolumn{4}{c}{\textbf{S3}}\\
     \cmidrule(l){2-5} \cmidrule(l){7-10} \cmidrule(l){12-15}
        &MRR  &@1   &@3   &@10                                       && MRR & @1  & @3  & @10  && MRR & @1  & @3  & @10\\
    \midrule
    DistMult            & 0.783 & 0.583 & 0.979 & \textbf{1.000} && 
    0.582 & \textbf{0.583} & 0.667 & \textbf{0.958} && 0.283 & 0.104 & 0.375 & 0.604 \\
    DistMult-\approach  & \textbf{0.990} & \textbf{0.979} & \textbf{1.000} & \textbf{1.000} && \textbf{0.753} & 0.458 & \textbf{0.938} & \textbf{0.958} && \textbf{0.670} & \textbf{0.542} & \textbf{0.771} & \textbf{0.896} \\
    \midrule
    Keci                & 0.716 & 0.479 & 0.917 & \textbf{1.000}  && \textbf{0.627} & 0.501 & 0.667 & 0.896 && 0.292 & 0.188 & 0.333 & 0.583 \\
    Keci-\approach      & \textbf{1.000} & \textbf{1.000} & \textbf{1.000} & \textbf{1.000} && 0.596 & \textbf{0.521} & \textbf{0.696} & \textbf{0.950} && \textbf{0.667} & \textbf{0.500} & \textbf{0.771} & \textbf{0.917} \\
    \midrule
    DeCaL               & 0.712 & 0.500 & 0.896 & 0.979 && 0.481 & 0.271 & 0.708 & \textbf{0.812} && 0.312 & 0.188 & 0.354 & 0.583 \\
    DeCaL-\approach     & \textbf{0.865} & \textbf{0.729} & \textbf{1.000} & \textbf{1.000} & & \textbf{0.685} & \textbf{0.546} & \textbf{0.812} & 0.562  && \textbf{0.590} & \textbf{0.417} & \textbf{0.729} & \textbf{0.875} \\
    \midrule
    MuRE                & 0.693        &0.458         & 0.979         & \textbf{1.000} && 0.677 & \textbf{0.458} & \textbf{0.896} & \textbf{0.979} && 0.205 & 0.021 & 0.333 & 0.500 \\
    MuRE-\approach      &\textbf{0.969}&\textbf{0.938}&\textbf{1.000} & \textbf{1.000} && \textbf{0.690} & 0.375 & 0.875 & \textbf{0.979} && \textbf{0.601} & \textbf{0.354} & \textbf{0.750} & \textbf{0.958} \\
    \midrule
    ComplEx             & 0.411 & 0.250 & 0.479 & 0.688 && \textbf{0.243} & \textbf{0.125} & \textbf{0.312} & \textbf{0.500 }&& 0.322 & 0.208 & 0.333 & 0.562 \\
    ComplEx-\approach   & \textbf{0.979} & \textbf{0.958} & \textbf{1.000} & \textbf{1.000} && 0.196 & 0.062 & 0.250 & \textbf{0.500} && \textbf{0.491} & \textbf{0.396} & \textbf{0.521} & \textbf{0.667} \\
    \midrule
    QMult              & 0.214 & 0.130 & 0.229 & 0.374 && \textbf{0.467} & \textbf{0.375} & 0.479 & 0.688 && 0.127 & 0.021 & 0.146 & 0.354 \\
    QMult-\approach    & \textbf{0.670} & \textbf{0.521} & \textbf{0.812} & \textbf{1.000} && \textbf{0.476} & \textbf{0.375} & \textbf{0.562} & \textbf{0.708} && \textbf{0.324} & \textbf{0.208} & \textbf{0.417} & \textbf{0.542} \\
    \midrule
    QuatE              & 0.384 & 0.188 & 0.500 & 0.708 
                        && 0.247 & \textbf{0.104}& 0.229 & 0.583 
                        && 0.103 & 0.042 & 0.062 & 0.208 \\
    QuatE-\approach    & \textbf{1.000} & \textbf{1.000} & \textbf{1.000} & \textbf{1.000} 
        && \textbf{0.250} & \textbf{0.104} & \textbf{0.271} & \textbf{0.604} 
        && \textbf{0.373} & \textbf{0.188} & \textbf{0.458} & \textbf{0.750} \\
     \bottomrule
    \end{tabular}
    \label{table:countries_enriched}
\end{table*}

\subsection{Numerical Literal Prediction}
\begin{figure}
\centering
\includegraphics[scale=0.50]
{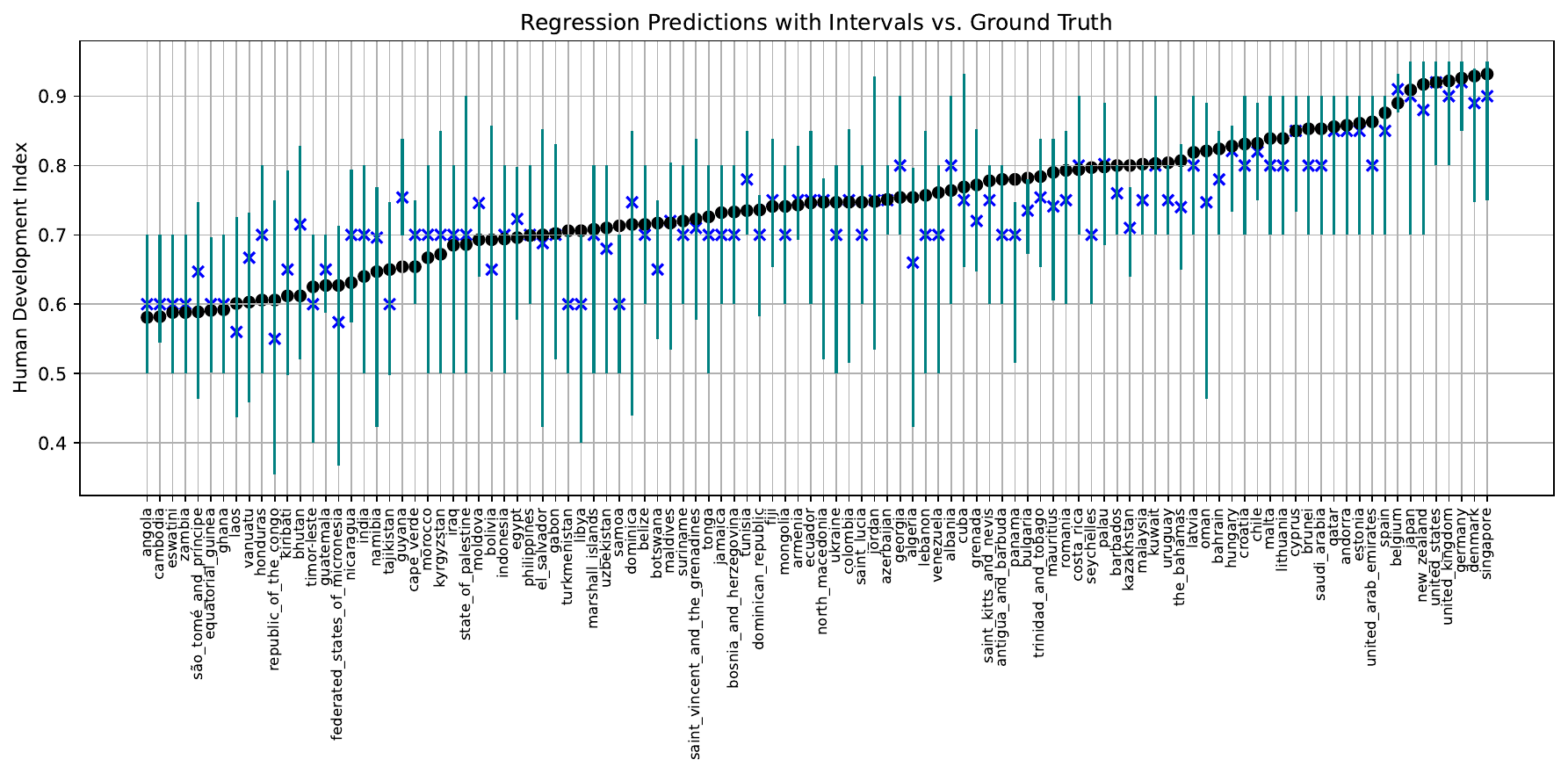}
\caption{RALP predictions for the human\_\allowbreak  development\_\allowbreak  index property (P1081) in LitWD1K dataset. Ground truth depicted in black dots, regression prediction in blue cross and interval range in teal lines.  For this property we report a mean squared error of 0.006 and a mean absolute error of 0.059. Some countries are omitted from the plot because of the limited space.}
\label{fig:preds_plot}
\end{figure}

Table \ref{table:numerical_lp_evaluation} presents the results of numerical literal link prediction on a selection of data properties from the \textit{LitWD1K} dataset, evaluated using the RALP framework. For each property, we report the true mean value ($y_{\text{avg}}$), standard deviation ($\sigma$), predicted mean value ($\hat{y}_{\text{avg}}$), ICR, and IW. 
The results indicate that the predicted mean values $\hat{y}_{\text{avg}}$ closely approximate the true means $y_{\text{avg}}$ across most properties. Additionally, the interval coverage rate—defined as the proportion of true values falling within the predicted bounds $[\hat{y}_{\text{min}}, \hat{y}_{\text{max}}]$—exceeds 75\% for all properties, indicating well calibrated interval estimates. At the same time, the reported interval widths (IW) suggest a reasonable trade-off between calibration and sharpness, with narrower intervals for more stable properties (e.g., \texttt{height\_metre}) and wider intervals for high-variance ones (e.g., \texttt{population}). Figure \ref{fig:preds_plot} further visualizes the regression predictions and their confidence intervals against ground truth values, illustrating the model’s capacity for calibrated uncertainty estimation.
To provide qualitative insight into the model’s reasoning, we include a representative example for the subject-relation pair 
$(s=\texttt{"gabon"}, \ r=\texttt{"nominal\_gdp\_per\_capita\_usd"}).$ 

The CoT module performs step-by-step reasoning as follows:
\begin{quote}
\textit{``Given the context, we know that Gabon's total nominal GDP is 14,622,880,885.6842 USD and its population is 2,025,137. To estimate the nominal GDP per capita, we can divide the total nominal GDP by the population. However, we also have data on other countries' nominal GDP per capita which can help us set a reasonable range. The lowest nominal GDP per capita among the listed countries is 499 USD (Sierra Leone) and the highest is 40,698 USD (United Arab Emirates). Given Gabon's economic status and the provided data, it is reasonable to assume that Gabon's nominal GDP per capita would fall somewhere between these two extremes.''}
\end{quote}
Based on this reasoning, the model predicts the following:
$\hat{y}_{\text{min}} = 1000.0$, $\hat{y} = 7225.0$, and $\hat{y}_{\text{max}} = 15000.0$.
The true value for this $(s, r)$ pair is $y = 7413.0$, which lies within the predicted interval, reflecting both the accuracy and the reliability of the model’s interval prediction capabilities. We want to note here that the main focus of numerical link prediction is the find a relatively close $\hat{y}$ to the true value $y$, whereas the interval prediction is made purely to ensure the reliability of the predicted values with no intended application.
\begin{table*}
    \caption{
    Numerical link prediction results in LitWD1K dataset. The shown data properties are selected randomly without replacement for the illustration purposes.
    We predict $\hat{y}$ together with interval bounds $\hat{y}_{\text{max}}$ and $\hat{y}_{\text{min}}$. For our evaluation we measure the mean value of ${y}$, standard deviation $\sigma$, the mean value of $\hat{y}$ and for the interval prediction, we measure interval coverage rate (ICR) for calibration and interval width (IW) for sharpness.}
    \centering
    \small
    \begin{tabular}{llllll}
    \toprule
     \textbf{Data Property} \ & $y_{\text{avg}}$ & $ \sigma$ \ & \ $\hat{y}_{\text{avg}}$ \ & \textbf{ICR} \ & \textbf{IW} \\
    \midrule
     population \ & \num{60923964} \ & $\pm$ 3\num{44077886} \ & \num{58504042} \ & 0.90 \ & \num{167280658} \ \\
     age\_of\_majority\_years\_old \ & 18.3 \ & $\pm$ 1.1 \ & 18.1 \ & 0.90 \ & 3.175 \ \\
     total\_fertility\_rate \ & 3.029 \ & $\pm$ 1.435 \ & 2.756 \ & 0.80 \ & 2.709 \ \\
     nominal\_gdp\_per\_capita\_usd \ & \num{11510} \ & $\pm$ \num{19408} \ & \num{11034} \ & 0.75 \ & \num{15140} \ \\
     human\_development\_index \ & 0.677 \ & $\pm$  0.145 \ & 0.698 \ & 0.79 \ & 0.236 \ \\
     life\_expectancy\_second \ & \num{2220348629} \ & $\pm$ \num{236968936} \ & \num{2204513934} \ & 0.81 \ & \num{640296028} \ \\
     real\_gdp\_growth\_rate \ & 0.025 \ & $\pm$ 0.036 \ & 0.028 \ & 0.78 \ & 0.114 \ \\
     mass\_kilogram \ & 68.7 \ & $\pm$ 7.4 \ & 69.9 \ & 0.80 \ & 19.5 \ \\
     height\_metre \ & 1.80 \ & $\pm$ 0.07 \ & 1.81 \ & 0.82 \ & 0.19 \ \\
     coordinate\_location\_longitude \ & 19.34 \ & $\pm$ 70.20 \ & 15.27 \ & 0.83 \ & 22.80 \ \\
     \bottomrule
    \end{tabular}
    \label{table:numerical_lp_evaluation}
\end{table*}

\subsection{OWL Reasoning/Instance Retrieval}
Table~\ref{tab:js_per_concept_type} presents a detailed analysis of model performance across various OWL concept types under different input configurations. Specifically, we evaluate the impact of two factors: (i) the syntactic representation of concept expressions (Manchester syntax vs. Description Logic (DL) syntax), and (ii) the format used for representing RDF triples in the knowledge graph (with or without namespace, i.e., using full IRIs or shortened identifiers).
The results show that the model achieves perfect performance on \textit{atomic} concepts across all configurations (Jaccard similarity = 1.000), underscoring its robustness for basic concept retrieval. The model also demonstrates strong performance on more complex constructs such as \textit{at least restriction}, \textit{conjunction}, \textit{disjunction}, and \textit{negation}, with the best results achieved when using Manchester syntax and full IRIs (1.000, 0.986, 0.984, and 0.887, respectively).

In contrast, performance degrades considerably for certain logical operators, particularly \textit{universal restrictions}, where the best configuration yields a Jaccard similarity of only 0.541. Similar challenges are observed for \textit{at most restrictions} and \textit{inverse properties}, suggesting that these constructs may pose greater difficulty for language model-based reasoning in instance retrieval settings.

Our ablation study further reveals that using Manchester syntax generally outperforms DL syntax, particularly when combined with full IRI representations. This configuration consistently led to superior results across most concept types. These findings suggest that Manchester syntax may align better with the language model’s pretraining, likely due to its more natural and human-readable structure. Additionally, representing entities and relations using full IRIs appears to provide essential contextual cues that facilitate more accurate reasoning.
\begin{table}
    \centering
    \caption{Results of instance retrieval on \textit{Father} datasets using RALP. We show the mean Jaccard similarity measured between the true set and the predicted set for each concept type where '\#' denotes the number of concepts per concept type, 'Namespace' denotes whether we included the namespace for triples in the graph or not, 'M' means that the expression was set in Manchester syntax and 'DL' means that the expression was set in Description Logics syntax. The best results for each concept type are marked in bold.}
    \begin{tabular}{@{}llcccc@{}}
        \toprule
        \textbf{Concept Types} & \textbf{\#} 
        & \multicolumn{2}{c}{\textbf{Namespace}} 
        & \multicolumn{2}{c}{\textbf{No Namespace}} \\
        \cmidrule(lr){3-4}  \cmidrule(lr){5-6}
         & & \textbf{M} \ & \textbf{DL} \ & \textbf{M} & \textbf{DL} \\
        \midrule
        Atomic              & 3 \ & \textbf{1.000 }\ & \textbf{1.000} \ & \textbf{1.000} & \textbf{1.000}    \\
        Negation   \           & 55 \  & \textbf{0.887 }\ & 0.803 \ & 0.869 & 0.801   \\
        Conjunction    \       & 19 \ & \textbf{0.986} \ & 0.881 \ & 0.973 & 0.973  \\
        Disjunction  \         & 33 \ & \textbf{0.984} \ & 0.775 \ & 0.969 & 0.818  \\
        Existential   \        & 44 \ & 0.656 \ & 0.700 \ & 0.673 & \textbf{0.764}   \\
        Universal     \        & 4 \ & \textbf{0.541} \  & 0.479 \ & 0.291 & 0.354    \\
        At least restriction \ & 12 \ & \textbf{1.000} \ & 0.900 \ & \textbf{1.000} & 0.900   \\
        At most restriction \  & 12 \ & 0.519 \ & \textbf{0.677} \ & 0.477 & 0.511  \\
        Nominals    \          & 40 \ & 0.646 \ & 0.670 \ & 0.600 & \textbf{0.740} \\
        Inverse     \          & 36 \ & 0.479 \ & 0.613 \ & 0.558 & \textbf{0.705}   \\
        \bottomrule
    \end{tabular}
    \label{tab:js_per_concept_type}
\end{table}

We also observe that including the syntax field $S$—a short natural language description of the logical concept—contributes positively to model performance. This field improves both the coherence and groundedness of the generated reasoning chains, supporting prior findings that emphasize the role of explicit contextualization in mitigating hallucinations and enhancing logical consistency in LLM outputs.

Overall, these results affirm the effectiveness of our method across a broad spectrum of logical constructs while highlighting specific areas, such as universal quantification and numerical restrictions, where targeted improvements could further enhance performance. Future work may also investigate the scalability of these methods to deeper nesting and more compositional concept expressions.

\section{Limitations}
One limitation we observed in the numerical link prediction setting arises when the contextual information provided to the LLM includes extreme outliers. In some cases, these outlier values deviate significantly from the distribution of other numerical values and can negatively impact the model's prediction quality. For instance, in the LitWD1K dataset, Venezuela's inflation rate is recorded as 16,988.4—an order of magnitude higher than the mean for other countries. When such outliers are included in the input context, they can distort the inferred value range for the interval prediction. This issue, however, can be mitigated through simple pre-processing techniques, such as filtering or down-weighting statistical outliers in the context.

Another practical constraint is that input context must be aligned with human-readable labels or descriptions, especially when entities or relations are referenced using identifiers (e.g., QIDs or URIs). Since LLMs are primarily trained on natural language, failing to map these identifiers to descriptive text reduces their ability to reason effectively over the input.

Lastly, our method relies on large pre-trained language models, which can present challenges for scalability in terms of inference time and computational resources. Token limitation into representing the context used for inference becomes an issue when dealing with large graphs. However, a solution to that would be to craft a sampling algorithm that selects the most relevant triples to go into the context. While our approach enables strong zero-shot performance and generalization, future work could explore ways to distill or compress the reasoning capabilities into lighter-weight models for broader deployment.

\section{Conclusion}

In this work, we introduced a novel paradigm for predicting missing links by redefining the traditional continuous vector-parameterized scoring function as a string-parameterized function, $\scoreFunc_{\beta^{\leq c}}$, leveraging the power of Large Language Models (LLMs) and automated prompt optimization. 
Unlike conventional Knowledge Graph Embedding (KGE) models that rely on static embeddings and struggle with unseen entities and relations, our approach learns a dynamic CoT prompt \stringparameters via few-shot examples and gradient-free optimization with MIPRO.
This string-based scoring function offers several key advantages: 
\begin{itemize}
    \item  it can be learned efficiently with a limited number of LLM calls,
    \item it inherently supports inference over triples involving previously unseen entities, relations, or literals by incorporating them directly into the prompt, and 
    \item the learned scorer can effectively augment training data for traditional KGE models, enhancing their generalization capabilities.
\end{itemize}
While sensitive to the syntactic representation of KG elements, our method demonstrates the viability of learning symbolic scoring functions using LLMs for KG tasks.
We presented two main approaches: \approach, which utilizes in-context learning with a default prompt, and \learnapproach, which employs MIPRO to learn an optimized prompt tailored to the specific KG and task. We also proposed a method (Algorithm \ref{alg:enrichment}) to leverage these approaches for enriching knowledge graphs by identifying potentially missing triples. Our framework also extends naturally to numerical literal prediction, where it achieves low mean squared error and mean absolute error—demonstrating strong predictive accuracy in this underexplored setting. Additionally, we applied our approach to OWL instance retrieval, showing that using Manchester syntax yields better LLM performance than traditional DL syntax for expressing class expressions. By redefining the scoring function and introducing automated prompt learning, our work opens a flexible and generalizable path toward robust, literal-aware knowledge graph completion—even in zero-shot or out-of-knowledge-base scenarios.

\bibliographystyle{unsrt}  
\bibliography{references}

@article{lewis2020retrieval,
  title={Retrieval-augmented generation for knowledge-intensive nlp tasks},
  author={Lewis, Patrick and Perez, Ethan and Piktus, Aleksandra and Petroni, Fabio and Karpukhin, Vladimir and Goyal, Naman and K{\"u}ttler, Heinrich and Lewis, Mike and Yih, Wen-tau and Rockt{\"a}schel, Tim and others},
  journal={Advances in neural information processing systems},
  volume={33},
  pages={9459--9474},
  year={2020}
}

@article{ram2023context,
  title={In-context retrieval-augmented language models},
  author={Ram, Ori and Levine, Yoav and Dalmedigos, Itay and Muhlgay, Dor and Shashua, Amnon and Leyton-Brown, Kevin and Shoham, Yoav},
  journal={Transactions of the Association for Computational Linguistics},
  volume={11},
  pages={1316--1331},
  year={2023},
  publisher={MIT Press One Broadway, 12th Floor, Cambridge, Massachusetts 02142, USA~…}
}

@inproceedings{petroni-etal-2021-kilt,
    title = "{KILT}: a Benchmark for Knowledge Intensive Language Tasks",
    author = {Petroni, Fabio  and
      Piktus, Aleksandra  and
      Fan, Angela  and
      Lewis, Patrick  and
      Yazdani, Majid  and
      De Cao, Nicola  and
      Thorne, James  and
      Jernite, Yacine  and
      Karpukhin, Vladimir  and
      Maillard, Jean  and
      Plachouras, Vassilis  and
      Rockt{\"a}schel, Tim  and
      Riedel, Sebastian},
    editor = "Toutanova, Kristina  and
      Rumshisky, Anna  and
      Zettlemoyer, Luke  and
      Hakkani-Tur, Dilek  and
      Beltagy, Iz  and
      Bethard, Steven  and
      Cotterell, Ryan  and
      Chakraborty, Tanmoy  and
      Zhou, Yichao",
    booktitle = "Proceedings of the 2021 Conference of the North American Chapter of the Association for Computational Linguistics: Human Language Technologies",
    month = jun,
    year = "2021",
    address = "Online",
    publisher = "Association for Computational Linguistics",
    url = "https://aclanthology.org/2021.naacl-main.200/",
    doi = "10.18653/v1/2021.naacl-main.200",
    pages = "2523--2544",
}

@inproceedings{opsahl-ong-etal-2024-optimizing,
    title = "Optimizing Instructions and Demonstrations for Multi-Stage Language Model Programs",
    author = "Opsahl-Ong, Krista  and
      Ryan, Michael J  and
      Purtell, Josh  and
      Broman, David  and
      Potts, Christopher  and
      Zaharia, Matei  and
      Khattab, Omar",
    editor = "Al-Onaizan, Yaser  and
      Bansal, Mohit  and
      Chen, Yun-Nung",
    booktitle = "Proceedings of the 2024 Conference on Empirical Methods in Natural Language Processing",
    month = nov,
    year = "2024",
    address = "Miami, Florida, USA",
    publisher = "Association for Computational Linguistics",
    url = "https://aclanthology.org/2024.emnlp-main.525/",
    doi = "10.18653/v1/2024.emnlp-main.525",
    pages = "9340--9366"
}

@inproceedings{zhou2022large,
  title={Large language models are human-level prompt engineers},
  author={Zhou, Yongchao and Muresanu, Andrei Ioan and Han, Ziwen and Paster, Keiran and Pitis, Silviu and Chan, Harris and Ba, Jimmy},
  booktitle={The Eleventh International Conference on Learning Representations},
  year={2022}
}

@article{liu2023pre,
  title={Pre-train, prompt, and predict: A systematic survey of prompting methods in natural language processing},
  author={Liu, Pengfei and Yuan, Weizhe and Fu, Jinlan and Jiang, Zhengbao and Hayashi, Hiroaki and Neubig, Graham},
  journal={ACM computing surveys},
  volume={55},
  number={9},
  pages={1--35},
  year={2023},
  publisher={ACM New York, NY}
}

@article{kojima2022large,
  title={Large language models are zero-shot reasoners},
  author={Kojima, Takeshi and Gu, Shixiang Shane and Reid, Machel and Matsuo, Yutaka and Iwasawa, Yusuke},
  journal={Advances in neural information processing systems},
  volume={35},
  pages={22199--22213},
  year={2022}
}

@inproceedings{
yang2024large,
title={Large Language Models as Optimizers},
author={Chengrun Yang and Xuezhi Wang and Yifeng Lu and Hanxiao Liu and Quoc V Le and Denny Zhou and Xinyun Chen},
booktitle={The Twelfth International Conference on Learning Representations},
year={2024},
url={https://openreview.net/forum?id=Bb4VGOWELI}
}

@inproceedings{
guo2024connecting,
title={Connecting Large Language Models with Evolutionary Algorithms Yields Powerful Prompt Optimizers},
author={Qingyan Guo and Rui Wang and Junliang Guo and Bei Li and Kaitao Song and Xu Tan and Guoqing Liu and Jiang Bian and Yujiu Yang},
booktitle={The Twelfth International Conference on Learning Representations},
year={2024},
url={https://openreview.net/forum?id=ZG3RaNIsO8}
}

@article{edge2024local,
  title={From local to global: A graph rag approach to query-focused summarization},
  author={Edge, Darren and Trinh, Ha and Cheng, Newman and Bradley, Joshua and Chao, Alex and Mody, Apurva and Truitt, Steven and Larson, Jonathan},
  journal={arXiv preprint arXiv:2404.16130},
  year={2024}
}

@article{he2024g,
  title={G-retriever: Retrieval-augmented generation for textual graph understanding and question answering},
  author={He, Xiaoxin and Tian, Yijun and Sun, Yifei and Chawla, Nitesh and Laurent, Thomas and LeCun, Yann and Bresson, Xavier and Hooi, Bryan},
  journal={Advances in Neural Information Processing Systems},
  volume={37},
  pages={132876--132907},
  year={2024}
}

@inproceedings{
khattab2024dspy,
title={{DSP}y: Compiling Declarative Language Model Calls into State-of-the-Art Pipelines},
author={Omar Khattab and Arnav Singhvi and Paridhi Maheshwari and Zhiyuan Zhang and Keshav Santhanam and Sri Vardhamanan A and Saiful Haq and Ashutosh Sharma and Thomas T. Joshi and Hanna Moazam and Heather Miller and Matei Zaharia and Christopher Potts},
booktitle={The Twelfth International Conference on Learning Representations},
year={2024},
url={https://openreview.net/forum?id=sY5N0zY5Od}
}

@inproceedings{demir2023clifford,
  title={Clifford embeddings--a generalized approach for embedding in normed algebras},
  author={Demir, Caglar and Ngonga Ngomo, Axel-Cyrille},
  booktitle={Joint European Conference on Machine Learning and Knowledge Discovery in Databases},
  pages={567--582},
  year={2023},
  organization={Springer}
}

@inproceedings{glorot2010understanding,
  title={Understanding the difficulty of training deep feedforward neural networks},
  author={Glorot, Xavier and Bengio, Yoshua},
  booktitle={Proceedings of the thirteenth international conference on artificial intelligence and statistics},
  pages={249--256},
  year={2010},
  organization={JMLR Workshop and Conference Proceedings}
}

@inproceedings{dettmers2018convolutional,
  title={Convolutional 2d knowledge graph embeddings},
  author={Dettmers, Tim and Minervini, Pasquale and Stenetorp, Pontus and Riedel, Sebastian},
  booktitle={Proceedings of the AAAI conference on artificial intelligence},
  volume={32},
  year={2018}
}

@article{dai2020survey,
  title={A survey on knowledge graph embedding: Approaches, applications and benchmarks},
  author={Dai, Yuanfei and Wang, Shiping and Xiong, Neal N and Guo, Wenzhong},
  journal={Electronics},
  volume={9},
  number={5},
  pages={750},
  year={2020},
  publisher={MDPI}
}

@article{demir2025ontolearn,
  title={Ontolearn---A Framework for Large-scale OWL Class Expression Learning in Python},
  author={Demir, Caglar and Baci, Alkid and Kouagou, N'Dah Jean and Sieger, Leonie Nora and Heindorf, Stefan and Bin, Simon and Bl{\"u}baum, Lukas and Bigerl, Alexander and Ngomo, Axel-Cyrille Ngonga},
  journal={Journal of Machine Learning Research},
  volume={26},
  number={63},
  pages={1--6},
  year={2025}
}

@article{demir2022hardware,
  title={Hardware-agnostic computation for large-scale knowledge graph embeddings},
  author={Demir, Caglar and Ngomo, Axel-Cyrille Ngonga},
  journal={Software Impacts},
  volume={13},
  pages={100377},
  year={2022},
  publisher={Elsevier}
}

@inproceedings{kwon2023efficient,
  title={Efficient Memory Management for Large Language Model Serving with PagedAttention},
  author={Woosuk Kwon and Zhuohan Li and Siyuan Zhuang and Ying Sheng and Lianmin Zheng and Cody Hao Yu and Joseph E. Gonzalez and Hao Zhang and Ion Stoica},
  booktitle={Proceedings of the ACM SIGOPS 29th Symposium on Operating Systems Principles},
  year={2023}
}

@misc{qwen2.5,
    title = {Qwen2.5: A Party of Foundation Models},
    url = {https://qwenlm.github.io/blog/qwen2.5/},
    author = {Qwen Team},
    month = {September},
    year = {2024}
}

@article{yang2024qwen2,
  title={Qwen2. 5 technical report},
  author={Yang, An and Yang, Baosong and Zhang, Beichen and Hui, Binyuan and Zheng, Bo and Yu, Bowen and Li, Chengyuan and Liu, Dayiheng and Huang, Fei and Wei, Haoran and others},
  journal={arXiv preprint arXiv:2412.15115},
  year={2024}
}

@article{qwen2,
      title={Qwen2 Technical Report}, 
      author={An Yang and Baosong Yang and Binyuan Hui and Bo Zheng and Bowen Yu and Chang Zhou and Chengpeng Li and Chengyuan Li and Dayiheng Liu and Fei Huang and Guanting Dong and Haoran Wei and Huan Lin and Jialong Tang and Jialin Wang and Jian Yang and Jianhong Tu and Jianwei Zhang and Jianxin Ma and Jin Xu and Jingren Zhou and Jinze Bai and Jinzheng He and Junyang Lin and Kai Dang and Keming Lu and Keqin Chen and Kexin Yang and Mei Li and Mingfeng Xue and Na Ni and Pei Zhang and Peng Wang and Ru Peng and Rui Men and Ruize Gao and Runji Lin and Shijie Wang and Shuai Bai and Sinan Tan and Tianhang Zhu and Tianhao Li and Tianyu Liu and Wenbin Ge and Xiaodong Deng and Xiaohuan Zhou and Xingzhang Ren and Xinyu Zhang and Xipin Wei and Xuancheng Ren and Yang Fan and Yang Yao and Yichang Zhang and Yu Wan and Yunfei Chu and Yuqiong Liu and Zeyu Cui and Zhenru Zhang and Zhihao Fan},
      journal={arXiv preprint arXiv:2407.10671},
      year={2024}
}

@article{hogan2021knowledge,
  title={Knowledge graphs},
  author={Hogan, Aidan and Blomqvist, Eva and Cochez, Michael and d’Amato, Claudia and Melo, Gerard de and Gutierrez, Claudio and Kirrane, Sabrina and Gayo, Jos{\'e} Emilio Labra and Navigli, Roberto and Neumaier, Sebastian and others},
  journal={ACM Computing Surveys (CSUR)},
  volume={54},
  number={4},
  pages={1--37},
  year={2021},
  publisher={ACM New York, NY, USA}
}

@inproceedings{ruffinelli2020you,
  author    = {Daniel Ruffinelli and
               Samuel Broscheit and
               Rainer Gemulla},
  title     = {You {CAN} Teach an Old Dog New Tricks! On Training Knowledge Graph
               Embeddings},
  booktitle = {8th International Conference on Learning Representations, {ICLR} 2020,
               Addis Ababa, Ethiopia, April 26-30, 2020},
  publisher = {OpenReview.net},
  year      = {2020},
  url       = {https://openreview.net/forum?id=BkxSmlBFvr},
  bibsource = {dblp computer science bibliography, https://dblp.org}
}

@dataset{gesese_2021_4701190,
  author       = {Gesese, Genet Asefa and
                  Alam, Mehwish and
                  Sack, Harald},
  title        = {LiterallyWikidata - A Benchmark for Knowledge
                   Graph Completion using Literals
                  },
  month        = apr,
  year         = 2021,
  publisher    = {Zenodo},
  version      = {0.1},
  doi          = {10.5281/zenodo.4701190},
  url          = {https://doi.org/10.5281/zenodo.4701190},
}

@inproceedings{baek-etal-2023-knowledge,
    title = "Knowledge-Augmented Language Model Prompting for Zero-Shot Knowledge Graph Question Answering",
    author = "Baek, Jinheon  and
      Aji, Alham Fikri  and
      Saffari, Amir",
    editor = "Dalvi Mishra, Bhavana  and
      Durrett, Greg  and
      Jansen, Peter  and
      Neves Ribeiro, Danilo  and
      Wei, Jason",
    booktitle = "Proceedings of the 1st Workshop on Natural Language Reasoning and Structured Explanations (NLRSE)",
    month = jun,
    year = "2023",
    address = "Toronto, Canada",
    publisher = "Association for Computational Linguistics",
    url = "https://aclanthology.org/2023.nlrse-1.7/",
    doi = "10.18653/v1/2023.nlrse-1.7",
    pages = "78--106"
}

@misc{guo2025evopromptconnectingllmsevolutionary,
      title={EvoPrompt: Connecting LLMs with Evolutionary Algorithms Yields Powerful Prompt Optimizers}, 
      author={Qingyan Guo and Rui Wang and Junliang Guo and Bei Li and Kaitao Song and Xu Tan and Guoqing Liu and Jiang Bian and Yujiu Yang},
      year={2025},
      eprint={2309.08532},
      archivePrefix={arXiv},
      primaryClass={cs.CL},
      url={https://arxiv.org/abs/2309.08532}, 
}

@misc{pryzant2023automaticpromptoptimizationgradient,
      title={Automatic Prompt Optimization with "Gradient Descent" and Beam Search}, 
      author={Reid Pryzant and Dan Iter and Jerry Li and Yin Tat Lee and Chenguang Zhu and Michael Zeng},
      year={2023},
      eprint={2305.03495},
      archivePrefix={arXiv},
      primaryClass={cs.CL},
      url={https://arxiv.org/abs/2305.03495}, 
}

@misc{hu2024enablingintelligentinteractionsagent,
      title={Enabling Intelligent Interactions between an Agent and an LLM: A Reinforcement Learning Approach}, 
      author={Bin Hu and Chenyang Zhao and Pu Zhang and Zihao Zhou and Yuanhang Yang and Zenglin Xu and Bin Liu},
      year={2024},
      eprint={2306.03604},
      archivePrefix={arXiv},
      primaryClass={cs.AI},
      url={https://arxiv.org/abs/2306.03604}, 
}

@misc{huang2022largelanguagemodelsselfimprove,
      title={Large Language Models Can Self-Improve}, 
      author={Jiaxin Huang and Shixiang Shane Gu and Le Hou and Yuexin Wu and Xuezhi Wang and Hongkun Yu and Jiawei Han},
      year={2022},
      eprint={2210.11610},
      archivePrefix={arXiv},
      primaryClass={cs.CL},
      url={https://arxiv.org/abs/2210.11610}, 
}

@article{youn2022kglm,
  title={Kglm: Integrating knowledge graph structure in language models for link prediction},
  author={Youn, Jason and Tagkopoulos, Ilias},
  journal={arXiv preprint arXiv:2211.02744},
  year={2022}
}

@article{he2024linkgpt,
  title={Linkgpt: Teaching large language models to predict missing links},
  author={He, Zhongmou and Zhu, Jing and Qian, Shengyi and Chai, Joyce and Koutra, Danai},
  journal={arXiv preprint arXiv:2406.04640},
  year={2024}
}

@article{shu2024knowledge,
  title={Knowledge graph large language model (KG-LLM) for link prediction},
  author={Shu, Dong and Chen, Tianle and Jin, Mingyu and Zhang, Chong and Du, Mengnan and Zhang, Yongfeng},
  journal={arXiv preprint arXiv:2403.07311},
  year={2024}
}

@article{geng2025prompting,
  title={Prompting disentangled embeddings for knowledge graph completion with pre-trained language model},
  author={Geng, Yuxia and Chen, Jiaoyan and Zeng, Yuhang and Chen, Zhuo and Zhang, Wen and Pan, Jeff Z and Wang, Yuxiang and Xu, Xiaoliang},
  journal={Expert Systems with Applications},
  volume={268},
  pages={126175},
  year={2025},
  publisher={Elsevier}
}

@article{DBLP:journals/corr/abs-1904-12575,
  author       = {Hongwei Wang and
                  Miao Zhao and
                  Xing Xie and
                  Wenjie Li and
                  Minyi Guo},
  title        = {Knowledge Graph Convolutional Networks for Recommender Systems},
  journal      = {CoRR},
  volume       = {abs/1904.12575},
  year         = {2019},
  url          = {http://arxiv.org/abs/1904.12575},
  eprinttype    = {arXiv},
  eprint       = {1904.12575},
  bibsource    = {dblp computer science bibliography, https://dblp.org}
}

@article{guo2020survey,
  title={A survey on knowledge graph-based recommender systems},
  author={Guo, Qingyu and Zhuang, Fuzhen and Qin, Chuan and Zhu, Hengshu and Xie, Xing and Xiong, Hui and He, Qing},
  journal={IEEE Transactions on Knowledge and Data Engineering},
  volume={34},
  number={8},
  pages={3549--3568},
  year={2020},
  publisher={IEEE}
}

@article{xiong2020approximate,
  title={Approximate nearest neighbor negative contrastive learning for dense text retrieval},
  author={Xiong, Lee and Xiong, Chenyan and Li, Ye and Tang, Kwok-Fung and Liu, Jialin and Bennett, Paul and Ahmed, Junaid and Overwijk, Arnold},
  journal={arXiv preprint arXiv:2007.00808},
  year={2020}
}

@article{wang2019knowledge,
  title={Knowledge graph construction and applications for web search and beyond},
  author={Wang, Peilu and Jiang, Hao and Xu, Jingfang and Zhang, Qi},
  journal={Data Intelligence},
  volume={1},
  number={4},
  pages={333--349},
  year={2019},
  publisher={MIT Press One Rogers Street, Cambridge, MA 02142-1209, USA journals-info~…}
}

\end{document}